 \documentclass[final,5p,times,twocolumn,authoryear]{elsarticle}

\usepackage{amssymb}
\usepackage{amsmath}
\usepackage{multirow}
\usepackage{subcaption}
\usepackage{lipsum}
\usepackage{hyperref}
\usepackage{stfloats}

\journal{Remote Sensing of Environment}

\begin{document}

\begin{frontmatter}



\title{Multimodal Urban Tree Detection from Satellite and Street-Level Imagery via Annotation-Efficient Deep Learning Strategies}


\author[first]{In Seon Kim}
\affiliation[first]{organization={Department of Computer Science, University of California, Davis},
            addressline={}, 
            city={Davis},
            postcode={95616}, 
            state={CA},
            country={USA}}

\author[aff1]{Ali Moghimi\corref{cor1}}
\ead{amoghimi@ucdavis.edu}

\cortext[cor1]{Corresponding author}

\affiliation[aff1]{
    organization={Department of Biological and Agricultural Engineering, University of California, Davis},
    city={Davis},
    state={CA},
    postcode={95616},
    country={USA}
}


\begin{abstract}
Beyond the immediate biophysical benefits, urban trees play a foundational role in environmental sustainability and disaster mitigation. 
Precise mapping of urban trees is essential for environmental monitoring, post-disaster assessment, and strengthening policy. 
However, the transition from traditional, labor-intensive field surveys to scalable automated systems remains limited by high annotation 
costs and poor generalization across diverse urban scenarios. This study introduces a multimodal framework that integrates high-resolution 
satellite imagery with ground-level Google Street View to enable scalable and detailed urban tree detection under limited-annotation conditions. 
The framework first leverages satellite imagery to localize tree candidates and then retrieves targeted ground-level views for detailed detection, 
significantly reducing inefficient street-level sampling. To address the annotation bottleneck, domain adaptation is used to transfer knowledge 
from an existing annotated dataset to a new region of interest. To further minimize human effort, we evaluated three learning strategies: 
semi-supervised learning, active learning, and a hybrid approach combining both, using a transformer-based detection model. 
The hybrid strategy achieved the best performance with an F1-score of 0.90, representing a 12\% improvement over the baseline model. 
In contrast, semi-supervised learning exhibited progressive performance degradation due to confirmation bias in pseudo-labeling, while active 
learning steadily improved results through targeted human intervention to label uncertain or incorrect predictions. Error analysis further 
showed that active and hybrid strategies reduced both false positives and false negatives. Our findings highlight the importance of a multimodal 
approach and guided annotation for scalable, annotation-efficient urban tree mapping to strengthen sustainable city planning.
\end{abstract}




\end{frontmatter}





\section{Introduction}
\label{sec:introduction}

In urban settings, trees provide a wide range of biophysical and social benefits that are critical to
both environmental sustainability and human well-being
\citep{Donovan2017}.
Their value has primarily been associated with measurable environmental functions, such as energy
savings, carbon sequestration, and air purification.
\citep{Donovan2017}
\citep{Peper2007}.
However, recent studies have highlighted the social importance of trees in urban areas. People often
develop emotional attachments to trees
\citep{Dwyer1991}
and green spaces, and access to these environments not only supports better physical health outcomes
but also promotes positive psychological well-being
\citep{Maas2009}.
In addition, street trees, as a key component of urban ecosystems, contribute significantly to
microclimate regulation and serve as an effective natural mechanism for mitigating urban heat island
effects
\citep{Li2018}.

To fully leverage the benefits of urban trees, high-resolution datasets that record the location,
condition, and structural attributes of trees are critically important. Accurate and consistent urban
tree mappings help environmental monitoring, including the assessment of post-disaster recovery
following events such as wildfires
\citep{Doerr2016}.
It also helps with urban tree canopy measurements, which could strengthen policy, budget
justification, and community engagement around tree planting and protection
\citep{Morgenroth2025}.
However, conventional approaches to tree data collection have been labor-intensive, inconsistent,
and limited in spatial and temporal scope, as they mostly rely on manual annotations of individual
trees. In densely built urban environments, the variability in tree composition and high spatial
heterogeneity of urban landscapes introduce site-specific differences that should be considered to
achieve highly accurate measurements
\citep{Zhao2018}.
As a result, there has been a growing shift from conventional urban tree mapping to remote sensing
and AI-powered approaches that can scale tree mapping, improve consistency across different domains,
and reduce dependency on manual collection.

To address the limitations of traditional field-based urban tree mapping, two observation strategies
have emerged: (1) overhead canopy assessment using high-resolution satellite imagery, and
(2) ground-level detection using street-level imagery. The availability of satellite data sources, such
as NAIP or commercial platforms (e.g., QuickBird), allowed studies to accurately localize and map
individual canopies
\citep{Ventura2024}
\citep{Arockiaraj2015}.
The inclusion of several spectral bands in high-resolution imagery provided finer spatial details,
supporting more robust species-level delineation
\citep{Lelong2020}.
However, despite the continental-scale coverage of these datasets
\citep{DEFRIES2007},
the complexity of heterogeneous urban scenes, with overlaps and shadows of tree canopies, reduced
detection accuracy
\citep{Zhang2022}.
Satellite imagery is also sensitive to seasonal changes, particularly when trees lack foliage, as the
absence of canopy structure in top-down imagery reduces the visual context needed to reliably detect
tree crowns
\citep{Arockiaraj2015}
\citep{Singh2005}.
The top-down view alone inherently lacks the three-dimensional structural information needed to
fully characterize urban trees
\citep{Zhang2022}.
More importantly, tree species identification from satellite imagery remains highly challenging due to
limitations in spatial resolution and the constraints imposed by the top-down viewing geometry.

In contrast to the broad coverage from satellite imagery, street-level data sources, such as Google
Street View (GSV), offer ground-level detail for characterizing trees. Unlike satellite imagery, GSV
provides a human-scale perspective of urban scenery with free and efficient access to large-scale
geospatial data
\citep{Biljecki2021}.
Studies have shown that street-level approaches outperform aerial methods in tree detection and
species identification
\citep{Wegner2016}.
However, ground-level detection struggles with scalability challenges. Occlusions caused by
obstacles that block the target tree remain a source of detection error
\citep{Choi2022}
\citep{Laumer2020}.
More importantly, street-level sampling is inefficient for creating large-scale urban tree inventories,
making city-wide analysis difficult to scale.

Although remote sensing has advanced urban tree mapping, single-modality methods still fail to
capture the full complexity of urban trees. While satellite imagery holds a distinct advantage for
localizing trees at scale due to its broad spatial coverage and scalability, it cannot capture the visual
details. Conversely, GSV offers human-scale detail of tree structure, but is vulnerable to occlusions
and scalability across large regions
\citep{Velasquez-Camacho2023}.

To translate image datasets captured by satellites or ground-level street views into actionable urban
tree information, studies increasingly relied on deep learning techniques for urban tree detection.
The use of machine learning resources further enables the understanding of street scene images
\citep{He2021},
in which various studies apply convolutional neural networks for tree detection (e.g., YOLOv3)
\citep{Choi2022}
or semantic segmentation (e.g., Mask R-CNN) to detect and locate separate tree instances
\citep{Lumnitz2021}.
These supervised learning methods require substantial annotation to achieve reasonable performance.
For instance,
\citet{Choi2022}
manually labeled nearly 50,000 individual tree instances across thousands of GSV images to achieve
a robust performance with 73\% accurate classification of tree species, while other studies utilized a
smaller dataset (e.g., 6,783 individual tree annotations), resulting in a detection accuracy of 56\% of
the total inventory
\citep{Laumer2020}.
As a result, a large, diverse annotated dataset is often necessary to achieve robust model
generalization. When trained on tree images from a particular region, models typically generalize
only to regions with comparable urban form and vegetation structure, and fail to scale across diverse
urban environments
\citep{Beery2022}.
This disconnect between the high cost of manual labeling and the need for models that could adapt
to diverse geographic contexts underscores the necessity for data-efficient learning strategies.

As deep learning approaches became more widely adopted, recent studies have highlighted the
limitations of directly transferring models trained in one visual domain to a different target domain.
Findings consistently indicate that full cross-domain generalization remains unreliable in the absence
of locally representative training data
\citep{Li2023}.
However, prior work shows that applying transfer learning and fine-tuning models pretrained on the
source domain with a limited number of target-specific samples can reduce domain transfer errors,
enabling performance levels closer to in-domain training
\citep{Li2023}.
While transfer learning is an effective strategy for leveraging existing knowledge across related tasks
or domains, the increasing complexity of modern deep learning models makes them susceptible to
overfitting when only small target-domain datasets are available
\citep{Liu2025}
\citep{Yu2023}.

The data dependency of contemporary deep learning models has led to a growing body of research
aimed at minimizing the annotation requirements, specifically in large-scale urban object mapping,
where manual labeling remains the primary bottleneck. Among these approaches, semi-supervised
learning (SSL) techniques leverage abundant unlabeled data to improve the model's representation
learning and generalization, even when labeled samples are minimal. Studies have explored various
strategies, such as noisy student training protocols, in which the teacher network (initially trained on
labeled data) generates pseudo-labels for unlabeled data that are then used to train a student network
(optimized using both labeled and pseudo-labeled data)
\citep{Xie2020},
and consistency regularization approaches that minimize the distance between a sample's predictions
and those of its augmented versions
\citep{Gao2020}.

Complementary to this, active learning (AL) seeks to strategically request human annotation only
for unlabeled samples predicted to be most informative, thereby optimizing labeling effort. Existing
AL algorithms are mainly divided into two categories: (1) representativeness sampling, selecting
samples that best represent the data distribution, and (2) uncertainty sampling, selecting samples the
current model is most uncertain about
\citep{Yang2015}.
Studies have shown that active learning methods demonstrate minimal loss in precision while
reducing the amount of human annotation to 20\%
\citep{Boukthir2022}.
Despite the advantage of reducing human labor in annotations, limitations exist. Called the cold start
problem, where insufficient initial labeled data leads to highly-biased models and poor subsequent
selection quality
\citep{Gao2020}
\citep{Konyushkova2017}
\citep{Houlsby2014}.
These limitations highlighted the need for more effective learning strategies that could improve
model performance while minimizing reliance on large labeled datasets.

Despite all recent advances, existing urban tree mapping approaches remain limited by their reliance
on single sensing modalities and annotation-intensive learning pipelines, which constrain scalability
and cross-domain generalization. To address these challenges, we propose a synergistic framework
that leverages the scalable spatial coverage of satellite imagery to guide detailed tree characterization
using ground-level imagery. By integrating multimodal sensing with a hybrid learning strategy that
combines transfer learning, semi-supervised learning, and active learning, our approach reduces
reliance on large-scale manual annotation while improving model generalization across domains.


\section{Methods}
\label{sec:methods}

\subsection{Overview of the Proposed Multimodal Aerial and Ground-Level Framework}
\label{sec:overview}

Urban tree mapping presents inherent challenges due to complex spatial structure, heterogeneous
visual contexts, and viewpoint-specific limitations in both aerial and ground-level imagery. Aerial
imagery offers consistent, large-scale coverage but limited structural detail, while ground-level
imagery provides fine-scale visual context with restricted spatial extent. Therefore, a
single-modality approach struggles to achieve both accuracy and coverage in urban tree inventories.

To address these limitations, we propose a coordinated multimodal framework that integrates
satellite imagery and GSV imagery. In the first stage, high-resolution satellite imagery is used to
detect tree locations within the study area. In the second stage, the spatial coordinates of detected
trees are used to retrieve corresponding GSV panoramas, enabling targeted ground-level inspection
of individual trees. This proposed approach mitigates the inefficiency of exhaustive street-level
sampling in GSV imagery, thereby substantially reducing computational and annotation costs.

\subsection{Satellite Model}
\label{sec:satellite_model}

The region of interest for this study is the Los Angeles area, specifically the zones affected by the
devastating wildfires that happened in January 2025. We selected this domain because the Palisades
and Eaton fires destroyed over 37,000 acres and destroyed over 16,000 structures
\citep{Guirguis2025},
urging for a need for rapid, high-resolution damage assessment. We acquired high-resolution aerial
imagery from Google Satellite images with a spatial resolution of 0.3 meters per pixel, along with
a building detection 
\href{https://universe.roboflow.com/prashant-v603s/building-detection-dn7vv-evesl}{dataset}
that provides canopy annotations in satellite view and was curated to remove any duplicated images.
The final dataset comprises 974 satellite images sourced from both the open-source repository and
the Google Satellite images to ensure a diverse range of environmental contexts and tree canopy
densities.

Table~\ref{tab:data_partition} shows how the data was partitioned into train, validation, and test
among two distinct datasets. The test dataset contains the majority of the target domain images to
ensure robust evaluation of the model's ability to detect canopies in dense urban scenarios. For the
object detection task, we employed the RF-DETR architecture. In our preliminary experiments,
RF-DETR demonstrated superior performance for in-depth object detection in the 0.3\,m resolution
imagery compared to other widely-used models, such as Faster R-CNN and YOLO. RF-DETR is
described in greater detail in Section~\ref{sec:transformer_model}.

\subsubsection{Tree Filtration}
\label{sec:tree_filtration}

We filtered the initial satellite detections to isolate trees suitable for ground-level analysis. As GSV
imagery is limited to the public right-of-way, trees located on private property or far from the road
edge are often occluded or lack high-resolution views. To address this issue, we overlaid the tree
coordinates detected from the satellite model onto the street network
\href{https://geohub.lacity.org/datasets/6b7e5c319b5543fcb35b8507c3b7e2bf_34/explore}{dataset}
with a common reference system. With a buffer around all road polylines as the spatial filter, we
preserved only the trees located within the buffer. This constraint represents the GSV camera path
to exclude distant or visually inaccessible trees. The filtered trees were subsequently passed to the
next step for ground-level inspection using GSV imagery.

\subsection{Ground-Level Model}
\label{sec:ground_level}

\subsubsection{Data Acquisition and Preparation}
\label{sec:data_acquisition}

To examine model robustness to domain transfer in ground-level tree detection, two distinct visual
domains were considered: an open-source Chinese urban tree image dataset
\citep{Yang2023}
and GSV imagery from our target Los Angeles urban area. The Chinese dataset consists of 3,949
high-resolution images of tree structure and canopy characteristics, which help stabilize
early-stage model training. Our target domain, street-level GSV images, is generated from a
satellite-based sampling model and reflects the visual characteristics of the desired environment.

The Chinese dataset shares morphological traits with the GSV, but we have also realized that the
two datasets do not share the same species across the two domains. This dual-domain setup enabled
us to assess the model's generalization under limited target-domain annotations via cross-domain
transfer, where knowledge learned from the source Chinese dataset is adapted to the Los Angeles
GSV target domain, which contains lower image resolution and different tree species.

\subsubsection{Data Partitioning}
\label{sec:data_partitioning}

We partitioned the composite dataset into training, validation, and testing sets. To ensure balanced
validation and test sets, we split the data using a histogram of tree counts per image to evenly
distribute detection densities across both datasets.

Most of the Chinese dataset was allocated to the training split, while the majority of GSV images
were reserved for validation and testing. This split reflects the differing roles of the two domains.
The Chinese dataset provides a large, high-quality source of labeled tree imagery that reduces the
need for extensive manual annotation in the target domain. In contrast, GSV images represent our
target deployment region, and prioritizing them for validation and testing allows us to directly
evaluate how well the model was able to generalize to the target visual characteristics. This ensures
that the performance metrics primarily reflect the target-domain accuracy rather than bias toward
the source domain.

\begin{table*}[htbp]
\caption{Dataset partitioning for satellite and ground-level imagery across source and target domains used for model training, validation, and testing.}
\label{tab:data_partition}
\begin{tabular*}{\textwidth}{@{\extracolsep{\fill}} l l c c c c}
\hline
Platform & Source/Target domain & Train Images & Validation Images & Test Images & Total Images \\
\hline
\multirow{3}{*}{Satellite Imagery}
    & Open-source Satellite (source) & 154   & 44    & 0     & 198   \\
    & Google Satellite      & 176   & 100   & 500   & 776   \\
    & In total              & 330   & 144   & 500   & 974   \\
\hline
\multirow{3}{*}{Ground-Level Imagery}
    & Chinese urban tree dataset (source)        & 3,509 & 440   & 440   & 4,389 \\
    & GSV Images            & 254   & 1,780 & 1,712 & 3,746 \\
    & In total              & 3,763 & 2,220 & 2,152 & 8,135 \\
\hline
\end{tabular*}
\end{table*}

\subsubsection{Preprocessing and Geometric Alignment of GSV Images}
\label{sec:preprocessing}

In the ground-level detection stage, we started with the output of the tree filtration step that
contains all coordinates of the trees in our region of interest, the middle of the LA region, where
it is mostly an urban area. For each input coordinate, we fetched the corresponding metadata using
the Google Static StreetView API to identify the precise location of the nearest available GSV
panorama coordinates.

To accurately center each GSV image on the target tree, we first converted both the tree and
panorama coordinates to Universal Transverse Mercator (UTM) to align their CRS. With UTM coordinate points, we were able to work in the Cartesian system to
calculate the distance between the panorama coordinates and the target tree using the Euclidean
distance equation:

\begin{equation}
    d = \sqrt{(t_e - p_e)^2 + (t_n - p_n)^2}
    \label{eq:euclidean}
\end{equation}

where $(p_e,\, p_n)$ and $(t_e,\, t_n)$ denote the UTM eastings and northings of the panorama
and tree, respectively. We excluded tree--panorama pairs with distances greater than 20 meters to
ensure a reliable and consistent visual representation of individual trees, as they reduce effective
spatial resolution and rather increase the likelihood of multiple trees appearing within a single
frame.

\begin{figure}[htbp]
    \centering
    \includegraphics[width=\columnwidth]{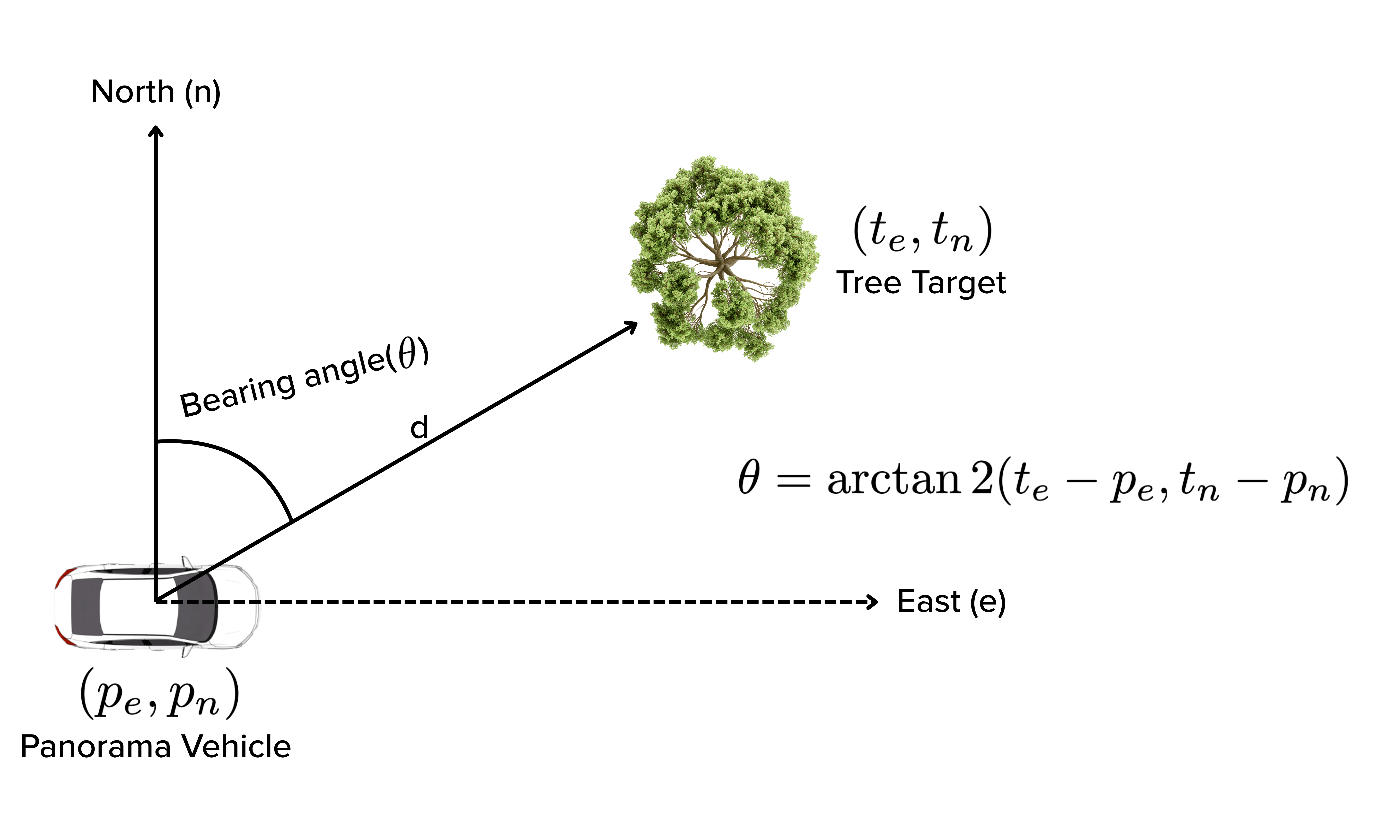}
    \caption{Illustration of the bearing angle calculation. The bearing angle $\theta$ from the
    panorama vehicle position $(p_e, p_n)$ to the tree target $(t_e, t_n)$ is computed in UTM
    coordinates, and passed as the bearing parameter to the Street
    View Static API to center the camera on the target tree.}
    \label{fig:heading_calc}
\end{figure}

For the bearing angle from the panorama to the tree coordinates, we used the arctangent formula. 
The resulting angle $\theta$ determines the heading parameter (Fig.~\ref{fig:heading_calc}),
which is then passed to the Street View API for image fetch. This ensures that the camera faces the
target tree directly.

\subsubsection{Transformer-Based Tree Detection Model}
\label{sec:transformer_model}

Unlike traditional detection models, such as Faster R-CNN or YOLO, the original DETR model
eliminates the need for hand-crafted anchor boxes and non-maximum suppression by employing
bipartite matching to align predictions with ground truth. However, the original DETR model had
several limitations, including slow convergence and difficulty detecting small objects due to weak
feature aggregation across scales
\citep{CarionNicolasandMassa2020}.

Various forms of DETR evolved to mitigate these limitations. Deformable DETR, which introduced
deformable cross-attention to focus computation effort on relevant spatial locations, improved
convergence speed and small-object detection
\citep{Zhu2021}.
However, the model remained complex and had computational overhead. Later, LW-DETR further
reduced the computational complexity by adopting the windowed and global attention mix, making
transformer-based detection more feasible for a real-time context
\citep{Chen2024}.

Building on these two architectures, we selected the RF-DETR model as our base model. It
incorporates the DINOv2 Vision Transformer backbone, which is pre-trained on large unlabeled
datasets. This backbone provides richer contextual representation and enhanced domain
adaptability
\citep{Sapkota2025},
making it well-suited for experiments in semi-supervised, active learning, and hybrid learning
scenarios where labeled data may be limited.

To justify the selection of the RF-DETR model, we benchmarked it against YOLOv11, a widely
used object detection framework, on the same dataset. As shown in
Table~\ref{tab:model_comparison}, RF-DETR demonstrated superior performance across all key
metrics, confirming its suitability for detecting trees in urban street-level environments. Recall
showed an extensive difference of 8.3\% improvement for the RF-DETR model. This is
particularly significant, as it demonstrates the ability to detect many more ground-truth trees in
scenarios where the trees are farther in the background, in complex settings, or with partially
occluded, irregularly shaped trees.

\begin{table}[htbp]
\caption{Model performance comparison for ground-level urban tree detection.}
\label{tab:model_comparison}
\begin{tabular}{l c c c c}
\hline
Model    & Precision & Recall & Best F1 & mAP50 \\
\hline
YOLOv11  & 0.85      & 0.67   & 0.75    & 0.79  \\
RF-DETR  & 0.86      & 0.75   & 0.80    & 0.85  \\
\hline
\end{tabular}
\end{table}

\subsubsection{Annotation-Efficient Learning Strategies}
\label{sec:learning_strategies}

\paragraph{Quantitative Evaluation of Learning Strategies}
\label{sec:quantitative_eval}

Large-scale urban tree mapping is constrained by the high cost and time required for manual
annotations, particularly for ground-level imagery where tree instances exhibit significant
variability and occlusion. To reduce reliance on large labeled datasets while maintaining detection
performance, we evaluated three learning strategies, including semi-supervised learning, active
learning, and hybrid learning. We designed these approaches to leverage abundant unlabeled data
and selectively incorporate human supervision, with the aim of effective model training starting
from a limited number of annotated samples.

\paragraph{Semi-Supervised Model}
\label{sec:ssl}

\begin{figure*}[!t]
    \centering
    \begin{subfigure}{0.48\textwidth}
        \centering
        \includegraphics[width=\linewidth]{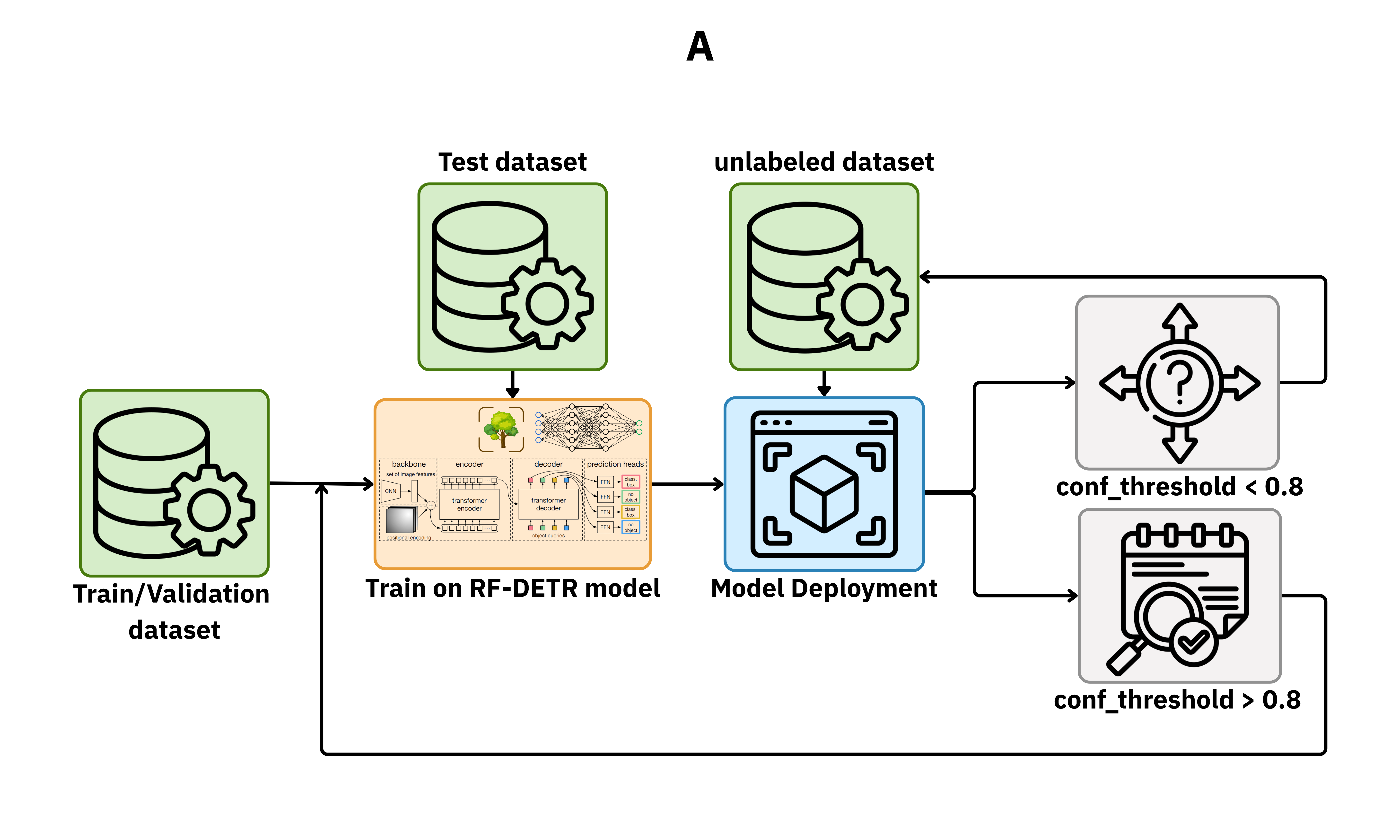}
        \label{fig:pipelines_a}
    \end{subfigure}
    \hfill
    \begin{subfigure}{0.48\textwidth}
        \centering
        \includegraphics[width=\linewidth]{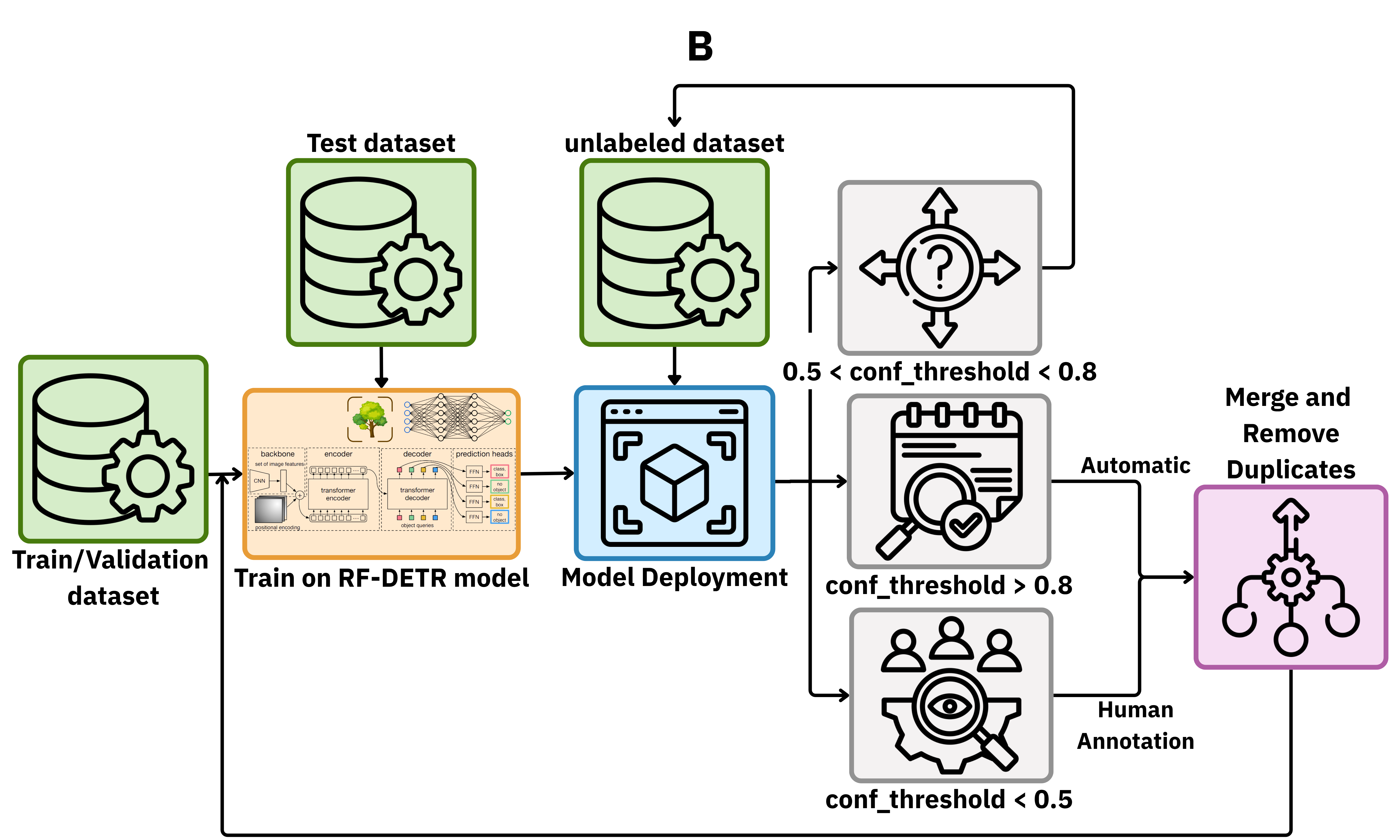}
        \label{fig:pipelines_b}
    \end{subfigure}
    \caption{Flowcharts for the three annotation-efficient learning strategies.
    (A) Flowchart for the semi-supervised pipeline. The model is trained on the labeled
    train/validation dataset and deployed on an unlabeled pool. Detections with confidence
    above 0.8 are automatically accepted as pseudo-labels and merged with the original dataset
    for retraining, while lower-confidence predictions are placed back in the unlabeled pool.
    (B) Active learning and hybrid learning strategies. In active learning, samples with
    prediction confidence below 0.5 are selected for human annotation. In the hybrid approach,
    predictions with confidence above 0.8 are automatically accepted as pseudo-labels
    (semi-supervised learning), while samples with confidence below 0.5 are assigned for human
    annotation (active learning) before merging and deduplication.}
    \label{fig:pipelines}
\end{figure*}

The semi-supervised learning framework (Fig.~\hyperref[fig:pipelines]{\ref*{fig:pipelines}A}) adopts a fully model-driven
pseudo-labeling strategy without human intervention. The RF-DETR model is initially trained on
the original, limited labeled dataset, then applied to a randomly selected subset of unlabeled
GSV images to generate predictions. To ensure annotation reliability, only detections with a
confidence score greater than 0.8 are retained. This threshold was determined through empirical
inspection. While predictions exceeding 0.8 consistently produced accurate and reliable tree
detections, lower-confidence predictions experienced greater noise and more false positives. Thus,
we treated these predictions as pseudo-ground truth data, as the model was highly confident that
the input aligned well with the learned feature representations. For images with multiple
annotations, only the high-confidence detections are preserved, while lower-confidence predictions
are discarded to minimize the noise in the pseudo-label data.

The filtered high-confidence detections are then appended to the pseudo-labeled training set, which
is combined with the original annotated dataset to retrain the model in subsequent learning rounds.

\paragraph{Active Learning}
\label{sec:al}

The active learning approach (Fig.~\hyperref[fig:pipelines]{\ref*{fig:pipelines}B}) similarly begins with the initially trained
model, which is deployed on randomly selected unlabeled GSV images to generate
preliminary predictions. Images containing at least one annotation with a confidence score below
0.5 are flagged for human review. Analysis of the confidence score distribution across the
unlabeled image pool, using a frequency histogram, indicates that predictions below the 0.5
threshold represent the hard examples where the model struggles the most to discriminate. Thus,
we targeted these low-confidence examples through human annotations to help maximize
information gain, forcing the model to explicitly learn the complex scenarios it could misinterpret.
During this human review, we discarded inaccurate or unusable predictions and refined or
confirmed the remaining detections. The resulting human-verified annotations are combined with
the original labeled dataset to retrain the model.

\paragraph{Active Learning + Semi-Supervised Model}
\label{sec:hybrid}

The hybrid learning approach also begins with the initially trained model
(Fig.~\hyperref[fig:pipelines]{\ref*{fig:pipelines}B}). The trained model is then deployed on randomly selected
unlabeled GSV images to generate preliminary predictions. From these predictions, two main
pathways are followed. Firstly, all images with prediction confidence greater than 0.8 are
automatically accepted as reliable detections. These annotations are initially saved as the
high-confidence dataset. Then, images containing at least one annotation with prediction
confidence below 0.5 are flagged for human review. Similar to the active learning strategy,
manual inspection cleaned the dataset, retaining only annotations that met quality and consistency.

After human reviewers revise the annotations requiring correction, the refined samples are merged
with fully verified annotations. Any images overlapping with the model's high-confidence
annotations are resolved by only keeping the human-annotated version. The combined datasets,
free of duplicates, are then used to retrain the RF-DETR model along with its initial datasets.

\section{Results and Discussion}
\label{sec:results}

\subsection{Model Performance on Satellite Images}
\label{sec:results_satellite}

\begin{figure}[htbp]
    \centering
    \includegraphics[width=\columnwidth]{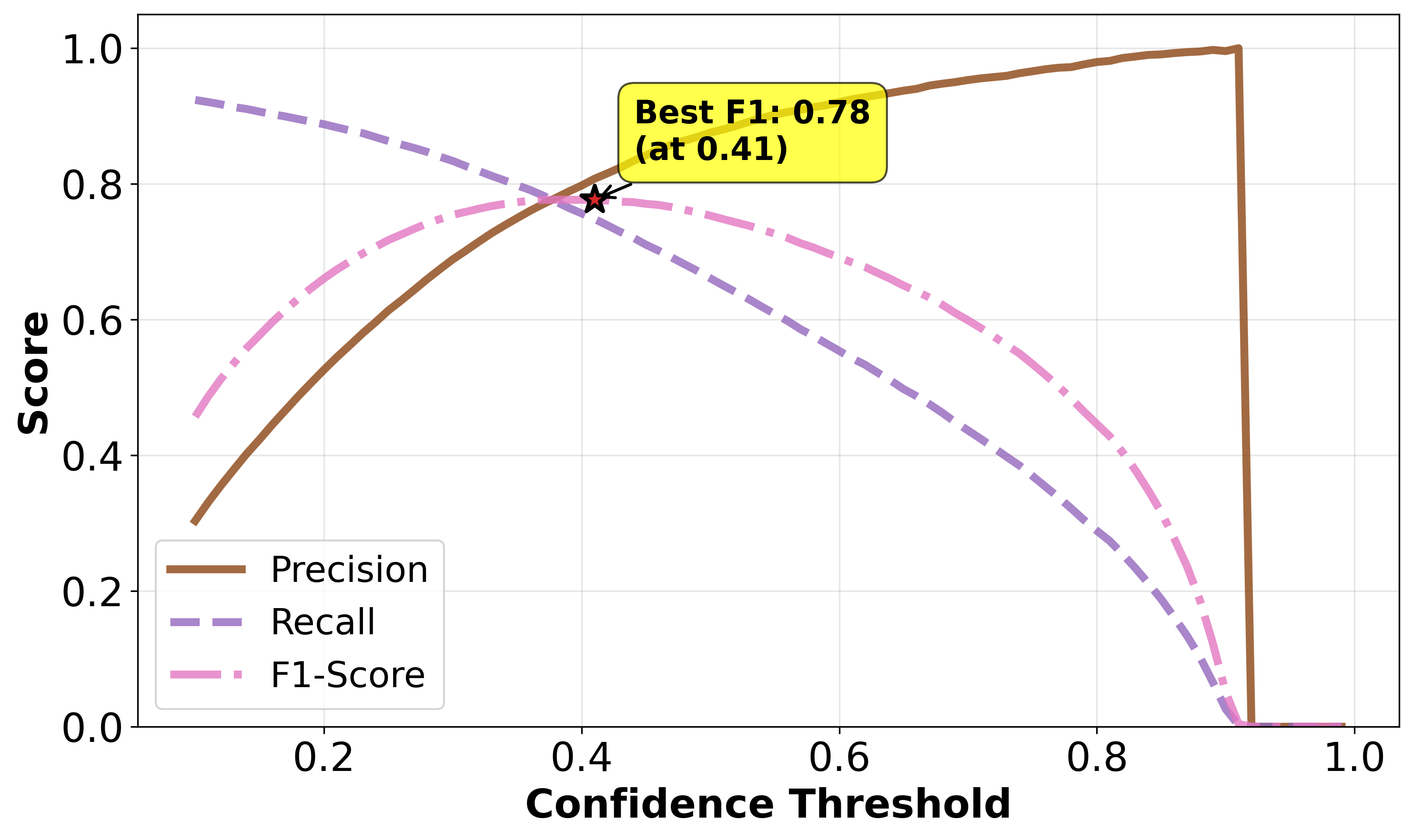}
    \caption{Precision, Recall, and F1-Score curve of the satellite model canopy detection. The best F1-score of 0.78 is achieved at a confidence threshold of 0.41. Lower thresholds increase recall at the cost of lower precision, while higher thresholds improve precision but reduce recall.}
    \label{fig:satellite_curve}
\end{figure}

As shown in Fig.~\ref{fig:satellite_curve}, the model trained on the open-source satellite 
\href{https://universe.roboflow.com/new-u5mtj/building-detection-dn7vv}{dataset}
and the target domain dataset (Google Satellite images of LA) achieved its highest F1-score of
0.78 at a confidence threshold of 0.41, along with a precision of 0.81 and a recall of 0.75. At
lower thresholds, the model includes a larger number of candidate detections, which increases
recall at the cost of more false positives. Conversely, a higher threshold filtered out uncertain
detections, improving precision at the expense of lower recall.

Despite the model's robustness, several factors likely contribute to the observed performance gap
of approximately 20\%. The model's ability to adapt from the source domain to the target domain
was largely constrained by data scarcity and domain-specific visual ambiguities. First, the limited
number of Google Satellite images used for training (e.g., 176 images) constrained the model's
exposure to the full range of visual variability present in the target environment. This limited the
model's ability to generalize across diverse canopy appearances and urban contexts. Second,
spectral similarity in residential environments, where lawns, yards, and other non-target vegetation
share spectral and structural characteristics with tree canopies, led to an increased rate of false
positives, especially in the absence of height information. Finally, the model struggled to detect
overlapping canopies as distinct instances, which complicated accurate detection in highly
clustered environments.

Despite these challenges, the RF-DETR model achieved competitive performance on the Google
Satellite test dataset with an F1-score of 0.78 (Table~\ref{tab:results_summary} and
Fig.~\ref{fig:satellite_curve}), which is consistent with similar large-scale urban tree detection
studies. For instance, a recent study on urban tree detection in California
\citep{Ventura2024}
reported an F1-score of 0.74 using high-resolution satellite imagery, trained on a dataset of 1,651
images, whereas our approach achieves competitive performance using 974 images
(\textit{i.e.}, ${\sim}$40\% reduction in annotation effort). Our model's F1-score slightly
exceeds the benchmark of this study due to the higher spatial resolution of our input data,
enabling the model to mitigate the ambiguity caused by small or dense, overlapping canopies,
which often limit performance.

\begin{table}[htbp]
\caption{Performance comparison of learning strategies on ground-level urban tree detection.}
\label{tab:results_summary}
\begin{tabular}{l c c c}
\hline
Model & Precision & Recall & F1-Score \\
\hline
Satellite Model            & 0.81 & 0.75 & 0.78 \\
Semi-Supervised (Avg)      & 0.85 & 0.70 & 0.77 \\
Active Learning (Avg)      & 0.89 & 0.84 & 0.86 \\
Hybrid Learning (Avg)      & 0.90 & 0.84 & 0.87 \\
\hline
\end{tabular}
\end{table}

\subsection{Model Performance on GSV Images}
\label{sec:results_gsv}

\subsubsection{Semi-Supervised Learning Results}
\label{sec:results_ssl}

Our semi-supervised approach relied solely on model-generated pseudo-labels derived from
high-confidence predictions, without any human intervention. While this learning strategy was
designed to minimize annotation effort by leveraging abundant unlabeled data, the results reveal
clear limitations in fully automated learning under complex urban conditions. The results showed
performance patterns distinct from those of the other two pipelines, with progressive degradation
across the 10 iterative rounds (Fig.~\ref{fig:learning_curves}). Precision remained relatively
stable, fluctuating between a relatively narrow range of 0.83 to 0.87 across all rounds. In
contrast, recall showed a consistent decreasing trajectory from 0.75 in round 1 to 0.67 in round
10, representing a 10.4\% decrease. This could be explained by the per-image annotation trend in
the semi-supervised pseudo-labeled dataset added at each training round. Most pseudo-labeled
images contained only a single high-confidence annotation; hence, the model preserved only the
most obvious, unoccluded instances of the trees while ignoring the harder instances, such as
partially occluded trees or those embedded in cluttered urban scenes. Since the model is retrained
on these sparse annotations, any valid tree that failed to meet the confidence threshold was
implicitly treated as an incorrect class, thereby penalizing the model for detecting valid trees.

In summary, this learning strategy forces the model to retain and learn only from high-confidence
detections, which predominantly correspond to visually obvious tree instances, while
systematically excluding more complex cases such as partially occluded, distant, or contextually
ambiguous trees.

The F1 score, similar to recall, demonstrated a decreasing pattern from 0.80 in round 1 to 0.76
in round 10. It also showed fluctuations in precision throughout the rounds. The maximum F1
score of 0.80 occurred in round 1, which is before any pseudo-labels were appended. This
indicates that the iterative accumulation of model-generated pseudo-labels, despite their high
confidence, introduces systematic biases that compound over training rounds and lead to
conservative detection behavior.

\subsubsection{Active Learning Results}
\label{sec:results_al}

Our active learning approach relied solely on the human-verified annotations. The overall metrics
of the model demonstrated consistent improvements across the 10 iterative rounds
(Fig.~\ref{fig:learning_curves}), which underscores the effectiveness of the active learning
strategy with selective human annotation. Precision showed an increasing pattern from 0.86 in
round 1 to 0.91 in round 10, demonstrating the most stable trajectory with less fluctuation among
the three learning strategies. Recall improved from 0.75 in round 1 to 0.87 in round 10, which is
a 16.2\% increase, and reached a plateau after round 7. The results reflect its learning of features
from human-corrected annotations across rounds. Lastly, the F1 score also increased in a steady
gain from 0.80 in round 1 to 0.89 in round 10. The consistent improvement of all metrics
indicates that selective human annotation of uncertain cases effectively addresses the model's
weaknesses and drives systematic performance enhancement.

\subsubsection{Active Learning + Semi-Supervised Learning Results}
\label{sec:results_hybrid}

The hybrid approach combined automated pseudo-labeling of high-confidence predictions with
targeted human annotation of low-confidence cases. This strategy was designed to balance
scalability and annotation efficiency while mitigating the confirmation bias observed in the purely
semi-supervised setting. This hybrid learning strategy demonstrated the most robust and
outstanding performance outcomes (Fig.~\ref{fig:learning_curves}). Precision showed initial
variability, dropping to 0.88 in round 3, but then recovered and stabilized, reaching 0.92 in round
nine. Recall experienced a substantial improvement from 0.75 in round 1 to peak values of 0.88,
which is around 17.4\% improvement. The F1 pattern reflected this balanced improvement,
initially starting with 0.80 in round 1 to its maximum value of 0.90 in round 10. While active
learning and the hybrid approach exhibited similar F1 performance, the hybrid strategy
consistently outperformed active learning across all 10 rounds.

\subsection{Training Convergence}
\label{sec:convergence}

\begin{figure*}[t]
    \centering
    \includegraphics[width=0.9\textwidth]{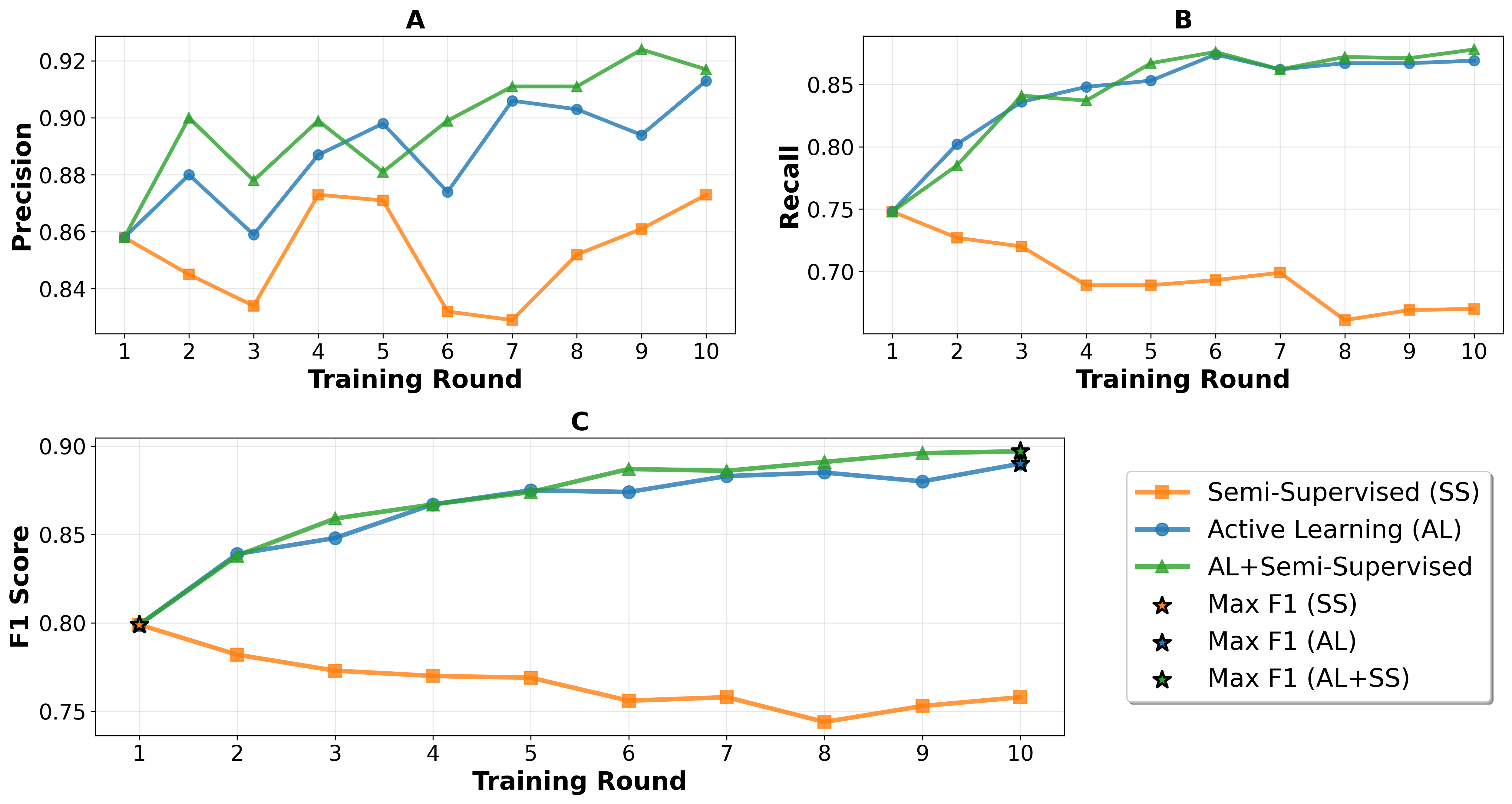}
    \caption{Performance comparison of three learning strategies over 10 rounds: Semi-Supervised
    learning (SS), Active Learning (AL), and hybrid AL + Semi-Supervised learning.
    (A) Precision trajectories across training rounds.
    (B) Recall trajectories across training rounds.
    (C) F1 score trajectories with peak performance markers (stars) for each method.}
    \label{fig:learning_curves}
\end{figure*}

We stopped our iterative training process at Round 10 based on the stabilization of our F1 score.
As shown in Fig.~\ref{fig:learning_curves}, while earlier rounds tend to yield a rapid change in
F1 score throughout three pipelines, the delta of the F1 score between rounds 9 and 10 was
statistically negligible, with a difference of less than 0.5\%. This plateau of the F1 score
indicates that the model has sufficiently learned and extracted meaningful information from the
sampling strategies and dataset. Continuing the strategy beyond this point could lead to a minimal
gain, but could increase the risk of overfitting to the specific biases of the pseudo-labeled data;
therefore, we stopped at round 10 in our study.

\section{Discussion}
\label{sec:discussion}

\subsection{Error Analysis and Performance Metric Dynamics}
\label{sec:error_analysis}

While precision, recall, and F1-score provide a concise summary of the overall performance, they
can obscure how different learning strategies affect the absolute number of correct and incorrect
detections. In our experiments, we observed that similar metric values could arise from
substantially different distributions of true positives, false positives, and false negatives. Directly
analyzing these detection counts offered more transparency in how each learning strategy altered
the model's detection behavior. It further enables us to view which strategy most effectively
increased correct detections while suppressing errors. Thus, to further examine these effects, we
inspected the underlying trends in true positives, false positives, and false negatives associated
with the metrics shown in Fig.~\ref{fig:detection_dynamics}.

\textbf{Semi-Supervised Learning.} Over successive semi-supervised training rounds, a systematic
degradation in the model's performance was observed, characterized by a decline in true positives
and a corresponding increase in false negatives. This behavior resulted in a consistent reduction
in recall and an increase in missed detections, as shown in Fig.~\ref{fig:detection_dynamics} and
Table~\ref{tab:detection_counts}. In the absence of external validation or human correction, the
model progressively reinforces its initial decision tendencies in previous rounds. If the initial
model exhibits conservative behavior, this tendency is amplified in subsequent rounds, leading the
model to increasingly suppress predictions on challenging instances. As a result, challenging tree
instances, such as partially occluded trees or those embedded in visually complex scenes, are less
likely to be detected by successive models, which led to the observed increase in false negatives.
This is because only high-confidence detections are retained as pseudo-labels, while missed trees
are never introduced into the training set, preventing the model from learning discriminative
features associated with these difficult cases.

Furthermore, as pseudo-labeled data accumulates across rounds, the effective decision boundary
shifts toward higher confidence thresholds. Therefore, tree instances that were marginally detected
in earlier rounds increasingly fall below the defined threshold in later iterations. This behavior is
reflected in the steady decline in true positives and the increase in false negatives. These
observations indicate that the semi-supervised learning strategy favors visually obvious tree
instances while progressively excluding more ambiguous or complex examples from the dataset
added to the next training cycle.

\textbf{Active Learning.} As summarized in Table~\ref{tab:detection_counts}, active learning
yields balanced and consistent improvements across all evaluation metrics. The number of true
positives increases while both false positives and false negatives decrease, indicating simultaneous
gains in detection sensitivity and error suppression. These trends suggest that the model becomes
progressively more accurate in detecting true tree instances while actively reducing incorrect
detections and missed trees.

By prioritizing low-confidence predictions for human review, the active learning strategy
systematically exposes the model to challenging examples that are underrepresented in randomly
sampled or pseudo-labeled datasets. Because human annotation is applied specifically to cases
where the model is uncertain, this approach directly addresses confirmation bias by correcting
errors rather than reinforcing existing model assumptions. In addition, when a low-confidence
detection is verified by a human as a true positive, the model learns to recognize similar tree
instances with greater confidence in subsequent rounds. Alternatively, when a human labels an
instance as a false positive, the model learns to suppress such misleading visual patterns.

Each training round introduces an average of 800 new training images of human-verified
annotations, resulting in steady performance gains (\textit{i.e.}, F1-score), as shown in
Fig.~\ref{fig:learning_curves}. This behavior indicates that human-in-the-loop corrections
effectively prevent the accumulation of systematic errors and enable controlled refinement of the
model's decision boundaries.

\textbf{Hybrid Learning.} The hybrid learning achieves the strongest performance as true positives
increase from 2,836 to 3,327, corresponding to a gain of 491 detections (approximately 17.31\%),
while false positives decrease from 469 to 303, representing a reduction of 166 incorrect
detections (approximately 35.39\%), and false negatives decrease from 955 to 464, indicating 491
fewer missed tree instances (approximately 51.41\%). The hybrid learning incorporates active
learning that identifies and corrects systematic errors along with semi-supervised learning that
propagates such corrections across similar unlabeled examples. The slight advantage of true
positives over pure active learning suggests that the hybrid approach also gains from the
semi-supervised component of successfully identifying and utilizing additional true positives that
active learning alone would require human annotation to capture. The two learning strategies
operate through complementary mechanisms: semi-supervised learning reinforces confident
predictions on easier tree instances, while active learning selectively targets ambiguous cases near
the decision boundary based on human correction.

\begin{table*}[tp]
\caption{Quantitative analysis of detection counts across learning strategies using true positives (TP), false positives (FP), and false negatives (FN). Percentages shown in parentheses indicate relative change from the initial round.}
\label{tab:detection_counts}
\begin{tabular*}{\textwidth}{@{\extracolsep{\fill}} l c c c c c c}
\hline
Learning   & Initial TP        & Final TP          & Initial FN & Final FN          & Initial FP & Final FP          \\
Strategy   & (Round 1)         & (\%)              & (Round 1)  & (\%)              & (Round 1)  & (\%)              \\
\hline
SS         & 2,836 (out of 3,791) & 2,538 ($-$10.5\%) & 955     & 1,253 ($+$31.2\%) & 469        & 369 ($-$21.3\%)   \\
AL         &                   & 3,295 ($+$16.2\%) &            & 496 ($-$48.1\%)   &            & 316 ($-$32.6\%)   \\
Hybrid     &                   & 3,327 ($+$17.3\%) &            & 464 ($-$51.4\%)   &            & 303 ($-$35.4\%)   \\
\hline
\end{tabular*}
\end{table*}

\subsubsection{True and False Detection Dynamics Across Iterative Learning}
\label{sec:detection_dynamics}

\begin{figure}[htbp]
    \centering
    \includegraphics[width=\columnwidth]{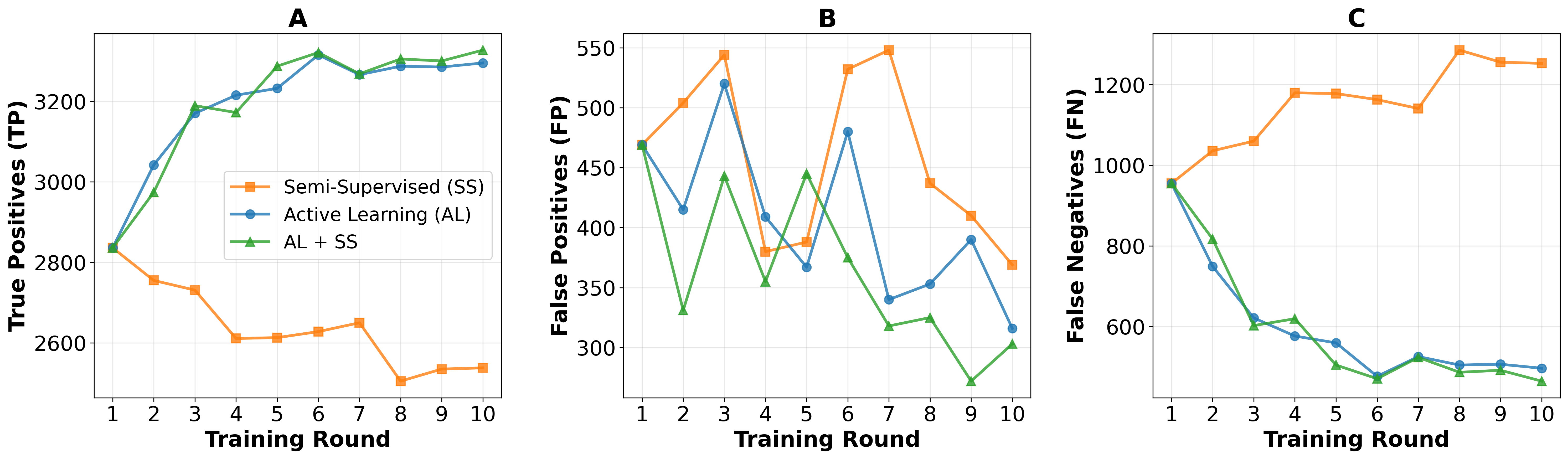}
    \caption{Error analysis and performance metric dynamics across the three learning strategies over iterative training rounds, including true positives (A), false positives (B), and false negatives (C).}
    \label{fig:detection_dynamics}
\end{figure}

For all three learning methods, Fig.~\ref{fig:detection_dynamics} illustrates higher fluctuations
in false positives across iterative rounds compared to other metrics. This is because false
positives arise from objects that look like trees but are not actually trees, so they lie near the
model's decision line of classification. As false positives are the borderline cases, even a small
addition of training images that slightly adjust confidence thresholds or update features could
change the outcome. This makes false positives unstable across training rounds. Sensitivity is
amplified for the learning pipelines as the model learns from its own pseudo-label predictions,
and human corrections dynamically reshape the decision boundary. Therefore, short-term
instability of false positives does not indicate the model's instability in the learning process, but
rather indicates the boundary refinement of the model.

\subsubsection{Sources of False Positive Errors}
\label{sec:fp_sources}

Our inspection of detection results revealed three primary sources of error that contributed to the
false positive rates observed across the three learning pipelines.

\paragraph{Duplicate Detections}
\label{sec:fp_duplicate}

\begin{figure}[htbp]
    \centering
    \includegraphics[width=\columnwidth]{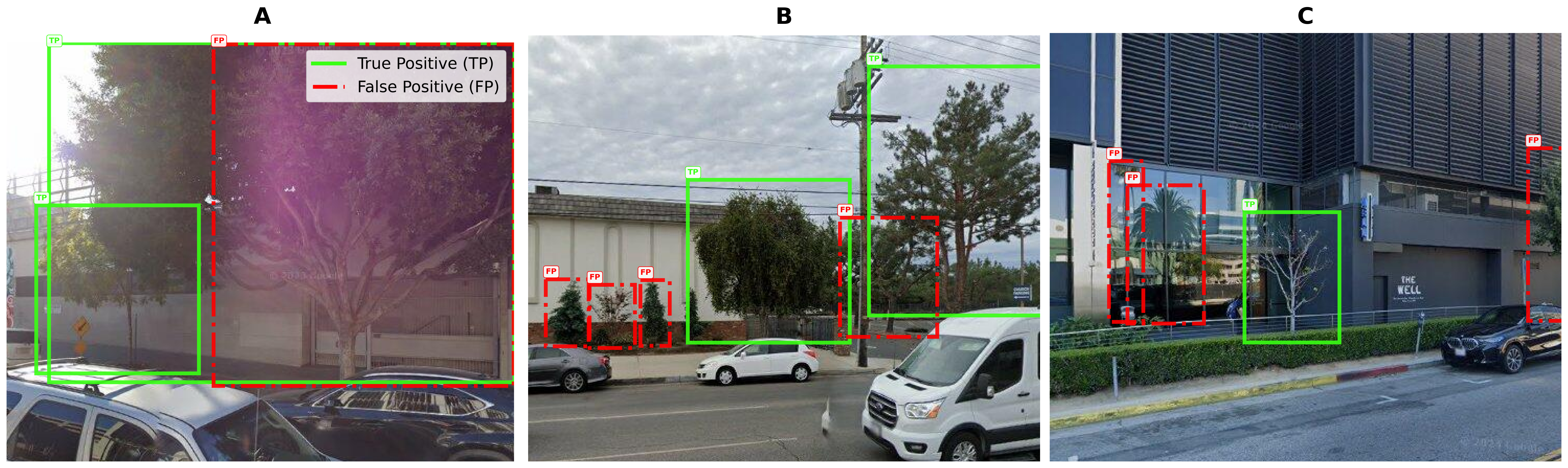}
    \caption{Error analysis of the tree detection model. (A) Multiple overlapping bounding boxes for a single tree instance due to the complex branching structure and occlusion. 
    (B) Model detecting trees accurately, but inaccurate human annotation of ambiguous cases leads to false positives. 
    (C) An urban scene with a modern building featuring a large reflective glass window; a nearby tree is correctly detected, while the model additionally identified the tree’s 
    reflection in the building’s windows as a separate tree instance, resulting in false positive detections driven by specular reflectance.
}
    \label{fig:error_analysis}
\end{figure}

Particularly in scenes with overlapping canopies, partial occlusions, or trees exhibiting multiple
stems emerging near the base of the trunk, the model correctly localized the tree but failed to
converge on a unified bounding box. Instead, it predicted multiple overlapping boxes for a single
tree (Fig.~\hyperref[fig:error_analysis]{\ref*{fig:error_analysis}A}).

\paragraph{Ground Truth Inconsistency}
\label{sec:fp_gt}

There were also apparent errors that could be attributed to human annotation variance rather than
model behavior. These cases typically arose from subjective labeling decisions in visually
ambiguous scenes, such as trees located in the background or those affected by occlusion and
unclear canopy boundaries (Fig.~\hyperref[fig:error_analysis]{\ref*{fig:error_analysis}B}).

\paragraph{Structural Ambiguity}
\label{sec:fp_structural}

Models also tended to misclassify vertical street structures, such as poles, as trees when a
separate tree canopy was present in the background (Fig.~\hyperref[fig:error_analysis]{\ref*{fig:error_analysis}C}).

\subsubsection{Analysis of High-Confidence False Positives}
\label{sec:hc_fp}

To understand the specific reasoning failures of the models, we manually inspected the highly
confident false positives, where the model assigned a confidence score greater than 0.7 to an
incorrect prediction. This analysis revealed that each learning pipeline exhibits distinct errors,
driven by its underlying training methods.

\paragraph{Cumulative Error in Semi-Supervised Learning}
\label{sec:hc_fp_ssl}

In semi-supervised learning, high-confidence errors exhibited a persistent and stable pattern,
meaning that non-tree objects consistently appeared as false positives across the ten rounds. Such
behavior is a direct consequence of the pseudo-labeling feedback loop in the absence of any
manual inspections. Once the model incorrectly identifies an object with high confidence, the
prediction is then converted to a ground truth label in subsequent rounds, which causes the error
to propagate through continued training. Without any human intervention or inspections to correct
the mistake, the model continually reinforces its own hallucination, effectively ``fossilizing'' the
error.

\paragraph{Human Annotation Bias Revealed by Active Learning}
\label{sec:hc_fp_al}

The active learning pipeline demonstrated a clear trajectory of error correction. Although the
number of high-confidence false positives appeared stable, manual inspection revealed that by
round 10, approximately 85\% of these cases were, in fact, true positives that had been overlooked
by human annotators. This annotation bias was present across all three models, but it became most
visible in the active learning framework due to its iterative uncertainty-driven sampling. The
model was able to effectively learn to generalize tree features, better than the inconsistency of
human annotators. It detected subtle or partially occluded instances that were overlooked during
the manual labeling process.

\paragraph{Hybrid Learning Effects on Detection Patterns}
\label{sec:hc_fp_hybrid}

The hybrid model showed distinct error results characterized by aggressive depth perception and
duplicate detections. One notable difference was that the hybrid model was able to detect distant
trees better than other pipelines. This stems from the amplification effect of the hybrid workflow.
Active learning teaches the model about small, distant tree features in the background, and the
semi-supervised learning understands the lesson and identifies more examples of similar distant
instances. As a result, hybrid learning performs in-depth detection, frequently identifying trees
located deeper in the scene that were not annotated by human labelers. Such detections were
counted as false positives during model evaluation, but manual inspection suggests that they often
correspond to true positives that were overlooked due to depth, scale, or visual ambiguity.

\subsection{Annotation Limitations}
\label{sec:annotation_limits}

Most of the studies that used GSV for urban tree mapping relied heavily on large manually
annotated datasets for supervised training
\citep{Velasquez-Camacho2023}.
The dependence on human annotation is time-consuming and costly, which limits the scalability
of ground-level tree detection approaches.

To address this limitation, our study introduced learning pipelines designed to strategically reduce
the need for human annotation while maintaining strong detection performance. Compared to prior
work using approximately 30,000 manually annotated ground-level trees to achieve 86.9\%
precision and 83.1\% recall
\citep{Velasquez-Camacho2023}, and recent continental-scale research that leveraged over 80,000
annotated images to achieve 92.2\% precision and 82.4\% recall \citep{Lake2026},
our approach achieved 85.8\% precision and 74.8\% recall with much less
training data size (9,366 initial annotations). After incorporating multiple learning pipelines and
expanding the annotation set to approximately 23,000 instances, performance further improved to
91.3\% precision and 87\% recall. This indicates that our approach was effective in reducing the
number of required annotations while maintaining or even improving the performance metrics.

\subsubsection{Domain Adaptation Gap}
\label{sec:domain_gap}

To further reduce the need for large-scale manual annotation, we employed domain adaptation to
leverage an existing annotated dataset (in this case, Chinese urban tree dataset) as a source
domain while targeting tree detection in downtown Los Angeles. However, transferring knowledge
between different domains introduced a challenge due to biological divergence and visual
differences between the datasets. We cross-referenced the species inventory of the Chinese dataset
with our Los Angeles region 
\href{https://geohub.lacity.org/datasets/lahub::trees-bureau-of-street-services/about}{dataset}
and found no species overlap. This divergence limits the direct transferability of specific tree
features learned from the source domain to our target domain, as specific bark textures, leaf
shapes, canopy shapes, etc., could be different.

Initial experiments using only the source dataset yielded suboptimal performance on our target
test dataset with a precision of 0.84, a recall of 0.73, and an F1-score of 0.78. To address this
domain gap, we integrated a subset of GSV imagery (Table~\ref{tab:data_partition}) into our
initial training set rather than relying exclusively on the cold-start transfer. This mixed-training
data yielded modest but consistent improvements across all metrics, achieving a precision of 0.86,
a recall of 0.75, and an F1-score of 0.80. The small subset of initial GSV images served as our
domain bridge for the model to slowly adapt to the visual features of our target domain trees
while substantially reducing the need for large-scale annotation with the target domain.

\subsection{Training Data Size Across Iterative Rounds}
\label{sec:training_data_size}

\begin{figure*}[tp]
    \centering
    \includegraphics[width=0.9\textwidth]{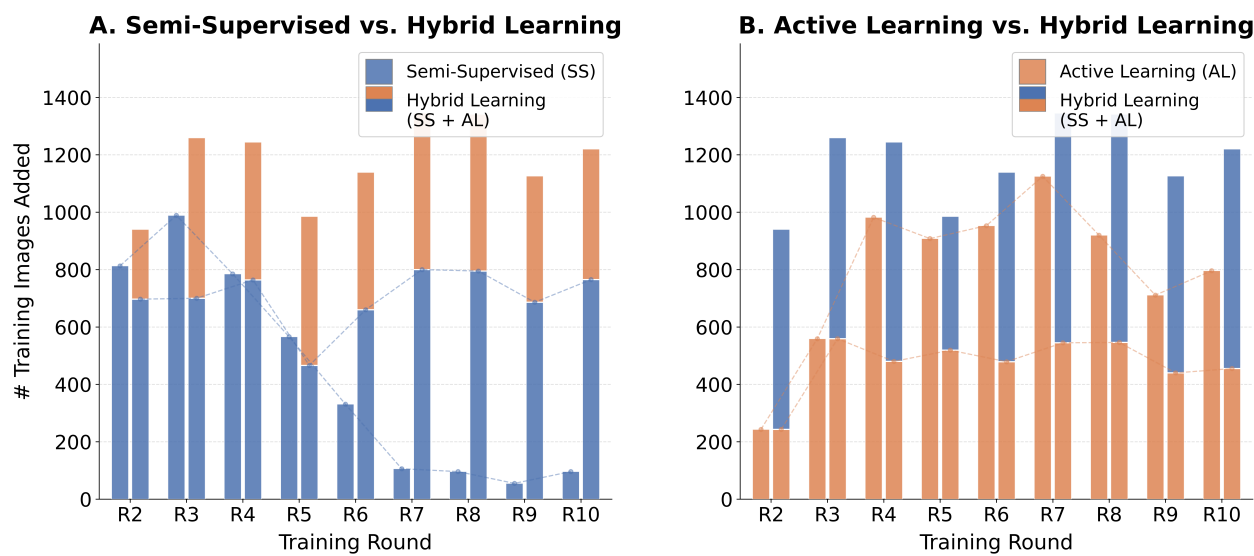}
    \caption{Distribution of the number of pseudo-labeled and human-annotated samples added to the training dataset per round for different learning strategies. 
    (A) Semi-supervised vs. hybrid learning: a decline in pseudo-label generation for the semi-supervised approach. 
    (B) Active learning vs. hybrid learning: active learning requires substantially higher human annotation effort, while the hybrid approach achieves comparable 
    dataset growth with fewer human annotations.}
    \label{fig:training_data}
\end{figure*}

The histogram in Fig.~\ref{fig:training_data} further illustrates the differences in annotation
labels of training data across learning strategies. As shown in Fig.~\hyperref[fig:training_data]{\ref*{fig:training_data}A}, the
number of pseudo-labeled training images added by the semi-supervised strategy shows a
noticeable decline after round 3, dropping from nearly 1,000 images to approximately 100 images
by round 10. This trend aligns with the performance degradation observed in
Fig.~\ref{fig:learning_curves} and reflects the model's increasingly conservative behavior
during iterative pseudo-labeling. Since only detections exceeding the confidence threshold (0.8)
were used as training labels, progressively fewer predictions satisfied this criterion as training
progressed and the model's prediction confidence declined. Consequently, the pseudo-label
progressively contracted, limiting the model's exposure to new training examples and reinforcing
the degradation in recall observed in later rounds (Fig.~\ref{fig:learning_curves}).

In contrast, the hybrid learning maintained a relatively stable number of high-confidence training
samples after round 5. This stability is because of integrating human-verified annotations in each
round, which continually introduces informative examples of challenging incidents. By combining
human corrections with automated pseudo-label propagation, the hybrid strategy sustains the
growth of the training dataset along with an increase in model performance.

Active learning exhibited a different pattern (Fig.~\hyperref[fig:training_data]{\ref*{fig:training_data}B}). While it
successfully added a large number of training images, it required substantially greater human
effort to annotate nearly 900 images by round 4. Compared to active learning, the hybrid learning
strategy achieved comparable dataset expansion while requiring significantly fewer human
annotations, demonstrating a more efficient balance between automated labeling and human
supervision.

\subsection{Advantages of the Multimodal Aerial--Ground-Level Framework}
\label{sec:multimodal_advantages}

Our multimodal framework provides several advantages over single-modality urban tree mapping.
Satellite imagery provides comprehensive spatial coverage for potential tree locations, whereas the
ground-level imagery captures detailed structural and contextual visual information that remains
inaccessible from the above-ground perspective. Our framework introduces a sequential
satellite-to-ground workflow in which satellite imagery guides the selective retrieval of
street-view tree coordinates. By pre-identifying precise tree locations, the system avoids
inefficient frame-by-frame image scanning and eliminates the need to iteratively test multiple
camera viewpoints, instead retrieving optimally positioned ground-level imagery directly at the
target coordinates. Such an advantage of multimodal integration is consistent with findings from
det2geo \citep{Wegner2016}, which demonstrated that
combining aerial and street-level imagery substantially outperformed single-modality detectors.


\section{Conclusion}
\label{sec:conclusion}

This study developed a multimodal urban tree detection framework that integrates high-resolution
satellite imagery and ground-level imagery to enable scalable and data-efficient urban tree
mapping. We addressed domain transfer challenges by incorporating a hybrid training dataset that
bridged source and target domains, mitigating confidence miscalibration and improving
cross-domain generalization. The satellite model achieved an F1-score of 0.78, along with the
ground-level hybrid learning pipeline reaching an average F1-score of 0.87, outperforming
semi-supervised learning and demonstrating up to 12.3\% improvement over initial performance
across iterative rounds. These results highlight the effectiveness of combining targeted human
annotation with automated pseudo-labeling to optimize performance under limited labeled data
conditions.

Although the proposed pipeline demonstrates robust performance, several structural challenges
remain. Occlusions and viewpoint constraints in ground-level imagery continue to cause missed
detections, particularly for partially hidden or distant trees. From the satellite perspective,
overlapping canopies and spectral similarity between trees and surrounding vegetation introduce
ambiguity that limits instance separation accuracy. In addition, the multimodal pipeline introduces
a dependency between stages, as ground-level tree detection relies on the accurate localization of
the satellite model. Therefore, errors in the satellite detection model may propagate to the
ground-level analysis, emphasizing the need to improve satellite detection accuracy in future work.

Beyond detection accuracy, the study highlights the broader potential of multimodal urban tree
mapping for large-scale environmental monitoring and urban ecosystem analysis. Accurate tree
localization and species identification can support applications such as urban forestry inventory
management, disaster mitigation, and city-scale environmental or ecological analysis. By
integrating different perspectives from satellite and ground-level imagery, multimodal learning
frameworks could help overcome the limitations of single-modality approaches and provide more
comprehensive representations of urban vegetation. This study demonstrates that strategically
guided multimodal learning is not only a methodological advancement but also a practical pathway
towards scalable and data-efficient urban tree mapping in complex urban environments.

\section*{CRediT authorship contribution statement}

\textbf{In Seon Kim:} Data curation, Formal analysis, Methodology, Validation, Visualization,
Writing -- original draft.
\textbf{Ali Moghimi:} Conceptualization, Funding acquisition, Methodology, Supervision,
Validation, Visualization, Writing -- review \& editing.

\section*{Acknowledgment}

This project is partially supported by the Western Center for Agricultural Health and Safety
(WCAHS) Grant Program funded under NIOSH grant \#U54OH007550. The authors would like to
express their sincere gratitude to Krishna Aggarwal and Grant Takeda for their invaluable
assistance with image annotation.

\bibliographystyle{elsarticle-harv}
\bibliography{export}






\end{document}